\documentclass{article}

\usepackage{microtype}
\usepackage{graphicx}
\usepackage{subcaption}
\usepackage{booktabs} 

\usepackage{hyperref}

\usepackage[accepted]{icml2026}

\usepackage{amsmath}
\usepackage{amssymb}
\usepackage{mathtools}
\usepackage{amsthm}

\usepackage[capitalize,noabbrev]{cleveref}

\theoremstyle{plain}

\theoremstyle{definition}

\theoremstyle{remark}

\usepackage[textsize=tiny]{todonotes}

\newcommand{\red}[1]{{\color{red}#1}}

\usepackage{tcolorbox}
\usepackage{colortbl}
\usepackage{xcolor}
\usepackage{amsmath}
\usepackage{algorithmic}
\usepackage{algorithm}
\usepackage{bm}
\usepackage{booktabs}
\usepackage{comment}
\usepackage{multirow}
\usepackage{framed}
\def\red#1{{\color{red} #1}}

\usepackage{url}
\makeatletter
\newcommand{\figcaption}[1]{\def\@captype{figure}\caption{#1}}
\newcommand{\tblcaption}[1]{\def\@captype{table}\caption{#1}}
\makeatother

\newcount\K
\def\bla#1{
\K=0 \loop\ifnum\K<#1
{\textcolor[gray]{0.9}{{\it bla bla bla bla bla bla bla bla bla bla bla bla bla bla bla}}}
\advance\K by1\repeat
}

\def\paragraph#1{\noindent \textbf{#1}}
\usepackage{colortbl,array,xcolor}
\usepackage{wrapfig}

\def\sref#1#2{\hyperref[#1]{\ref*{#1}#2}}
\usepackage{amsthm}
\usepackage{subcaption}

\usepackage[normalem]{ulem}



\usepackage{makecell}
\usepackage{dsfont}

\usepackage{amsmath,amsfonts,bm}

\def\eqref#1{equation~\ref{#1}}

\def\1{\bm{1}}

\DeclareMathAlphabet{\mathsfit}{\encodingdefault}{\sfdefault}{m}{sl}
\SetMathAlphabet{\mathsfit}{bold}{\encodingdefault}{\sfdefault}{bx}{n}

\DeclareMathOperator*{\argmax}{arg\,max}
\DeclareMathOperator*{\argmin}{arg\,min}

\newif\iftwollms
\twollmsfalse

\icmltitlerunning{Autoregressive Direct Preference Optimization}

\begin{document}

\twocolumn[
  \icmltitle{Autoregressive Direct Preference Optimization}



  \icmlsetsymbol{equal}{*}

  \begin{icmlauthorlist}
    \icmlauthor{Masanari Oi}{isct}
    \icmlauthor{Mahiro Ukai}{isct}
    \icmlauthor{Masahiro Kaneko}{mbzuai,isct}
    \icmlauthor{Naoaki Okazaki}{isct,aist,nii}
    \icmlauthor{Nakamasa Inoue}{isct}
  \end{icmlauthorlist}

  \icmlaffiliation{isct}{Department of Computer Science, Institute of Science Tokyo, Japan}
  \icmlaffiliation{aist}{AIRC, National Institute of Advanced Industrial Science and Technology (AIST), Japan}
  \icmlaffiliation{nii}{LLMC, National Institute of Informatics (NII), Japan}
  \icmlaffiliation{mbzuai}{Mohamed bin Zayed University of Artificial Intelligence, UAE}

  \icmlcorrespondingauthor{Masanari Oi}{masanari.oi@nlp.comp.isct.ac.jp}

  \icmlkeywords{Machine Learning, ICML, Reinforcement Learning, Large Language Models, Direct Preference Optimization}

  \vskip 0.3in
]



\printAffiliationsAndNotice{}  

\begin{abstract}
Direct preference optimization (DPO) has emerged as a promising approach for aligning large language models (LLMs) with human preferences.
However, the widespread reliance on the response-level Bradley-Terry (BT) model may limit its full potential, as the reference and learnable models are assumed to be autoregressive only after deriving the objective function.
Motivated by this limitation, we revisit the theoretical foundations of DPO and propose a novel formulation that explicitly introduces the autoregressive assumption prior to applying the BT model.
By reformulating and extending DPO, we derive a novel variant, termed \textbf{Autoregressive DPO (ADPO)}, that explicitly integrates autoregressive modeling into the preference optimization framework.
Without violating the theoretical foundations, the derived loss takes an elegant form: it shifts the summation operation in the DPO objective outside the log-sigmoid function.
Furthermore, through theoretical analysis of ADPO, we show that there exist two length measures to be considered when designing DPO-based algorithms: the token length $\mu$ and the feedback length $\mu'$.
To the best of our knowledge, we are the first to explicitly distinguish these two measures and analyze their implications for preference optimization in LLMs.
Project page: \url{https://stjohn2007.github.io/ADPO_project/}
\end{abstract}

\begin{table*}[t]
\centering
\small
\caption{\textbf{DPO versus ADPO.} By defining reward-based energy functions over the prefix closure $\mathcal{Y}^{*}$ of the output space $\mathcal{Y}$, we introduce the prefix-wise Bradley-Terry model. The derived loss shifts the summation operation outside the log-sigmoid function.
}
\setlength{\tabcolsep}{3pt}
\begin{tabular}{l|c|c|cccc}
\toprule
\textbf{Method} & \textbf{Energy domain} & \textbf{Modeling} & \textbf{Loss}\\
\midrule
DPO & Output space $\mathcal{Y}$ & Bradley-Terry & 
$- 
\log \sigma
\Bigl(
\beta \sum_{i} \log \frac{\pi_{\theta}(y^{w}_{i}| y^{w}_{<i}, x)}{\pi_{\mathrm{ref}}(y^{w}_{i}| y^{w}_{<i}, x)}
-
\beta \sum_{i} \log \frac{\pi_{\theta}(y^{l}_{i}| y^{l}_{<i}, x)}{\pi_{\mathrm{ref}}(y^{l}_{i}| y^{l}_{<i}, x)}
\Bigr)
$\\
\midrule
\makecell{ADPO\\ (Ours)}&
Prefix closure
$\mathcal{Y}^{*}$ &
\makecell{Prefix-wise \\ Bradley-Terry} &
$- 
\sum_{i} \log \sigma
\Bigl(
\beta \log \frac{\pi_{\theta}(y^{w}_{i}| y^{w}_{<i}, x)}{\pi_{\mathrm{ref}}(y^{w}_{i}| y^{w}_{<i}, x)}
-
\beta \log \frac{\pi_{\theta}(y^{l}_{i}| y^{l}_{<i}, x)}{\pi_{\mathrm{ref}}(y^{l}_{i}| y^{l}_{<i}, x)}
\Bigr)
$\\
\bottomrule
\end{tabular}
\label{tab:comparison}
\end{table*}

\section{Introduction}

Reinforcement learning from human feedback (RLHF) has emerged as a powerful paradigm for aligning large language models (LLMs) with human preferences~\citep{christiano2017deep, stiennon2020learning, ouyang2022instructgpt}.
As an alternative to traditional RLHF, Direct Preference Optimization (DPO)~\citep{rafailov2023DPO} analytically derives the optimal solution to the reward maximization objective, enabling efficient and scalable training.
This serves as a foundation for state-of-the-art LLMs, supporting diverse tasks ranging from conversational interactions to complex mathematical reasoning~\citep{zhang2024CPO, pang2024IRPO,jiao2025preference,liao2025tpo, lu2025stepcontrolled, shao2025tdpoDecay}.

Recent studies have introduced numerous variants of DPO primarily aimed at improving performance of LLMs~\citep{pang2024IRPO, zhao2024GPO, xiao2024CalDPO, meng2024SimPO, zhang2024CPO, chen2024SDPO, wu2024BetaDPO,  gupta2025AMPO, lai2025StepDPO, li2025TPO, wu2025AlphaDPO, son2025NSDPO, hu2025EXPO, chen2025LongPO, liang2025ROPO, yoon25confpo}.
However, most of them, including those proposing token-level objective functions~\citep{zeng24tdpo, zhu25tgdpo, lin2025critical}, fundamentally rely on the Bradley-Terry (BT) model~\citep{bradley1952rank},
which stipulates the human preference distribution at the level of complete responses.
In this modeling, there is an underlying assumption arising from RLHF that the reward function $r(x,y)$ is defined over complete responses $y \in \mathcal{Y}$, each paired with an input $x \in \mathcal{X}$.

This assumption is regarded as nearly essential because reward models typically learn to evaluate each input-output pair $(x, y)$.
For LLM training, even though models produce autoregressive distributions $\pi(y_{i}|y_{<i}, x)$, learning reward models at a finer granularity (\textit{e.g.}, at the token level) is challenging, as it is unrealistic to ask users to evaluate pairs $(x, y_{i})$ conditioned on incomplete responses $y_{< i}$.
However, since DPO does not require explicit reward models, there is an opportunity to introduce a more structured implicit reward function that better aligns with autoregressive modeling.

In this paper, we investigate how the autoregressive assumption can be explicitly incorporated into DPO when applying the BT model.
Specifically, we revisit the theoretical foundations of DPO and introduce a novel autoregressive variant, which we call \textbf{Autoregressive DPO (ADPO)}.
The core idea behind ADPO lies in the introduction of an implicit reward function $r^{*}(x, y_{\leq i})$ defined over the prefix closure $\mathcal{Y}^{*}$, a set of incomplete responses.
ADPO is then naturally derived through (1) reward-based prefix energies defined with an autoregressive reference model, (2) their corresponding Boltzmann distributions, and (3) the prefix-wise BT model.
The resulting loss function has an elegant form:
the summation operation in the DPO loss is shifted outside the log-sigmoid function, as summarized in Table~\ref{tab:comparison}.
To the best of our knowledge, we are the first to formulate and analyze such an autoregressive extension.
It is also worth noting that ADPO does not violate the theoretical foundations of DPO, as the difference naturally arises from the definition of the energy functions. In summary, our contributions are:
{
\setlength{\leftmargini}{14pt}
\begin{enumerate}
\setlength{\itemsep}{2pt}
\setlength{\parskip}{1pt}
\item[1)] 
We introduce \textbf{ADPO}, a novel DPO variant derived by extending the domain of energy from 
the output space $\mathcal{Y}$ to its prefix closure $\mathcal{Y}^{*}$. This enables the incorporation of the autoregressive assumption into DPO when applying the BT model, providing a theoretically consistent way to facilitate finer-granularity training.

\item[2)] We prove under mild assumptions that any reward function can be reparameterized by \textbf{an autoregressive model (Theorem 1)}.
Furthermore, through theoretical analysis, we identify two sequence length measures on $\mathcal{Y}$ that are critical for designing DPO-based algorithms: \textit{the token length measure} $\mu$ and \textit{the feedback length measure} $\mu'$. DPO corresponds to the special case of ADPO with $\mu'(y) = 1~(\forall y)$ (Corollary 1), while setting $\mu' \equiv \mu$ yields a token-level variant.
In general, choosing these two measures independently enables training at arbitrary granularity.
\end{enumerate}
}

\section{Related Work}

\paragraph{Preference Optimization.}
RLHF frames alignment as optimizing a reward based on pairwise human preferences~\citep{christiano2017deep, stiennon2020learning, ouyang2022instructgpt, xiong2024rlhfkl, ye2024oirlhf, zhu2023prlhf}. Online alignment methods maximize expected return under on-policy rollouts by employing Kullback--Leibler (KL) regularization techniques, such as trust-region constraints or explicit penalties relative to a reference policy~\citep{schulman2015trpo, schulman2017ppo, ouyang2022instructgpt}. For LLM training, PPO-based methods have explored improved trade-offs between return and divergence~\citep{wu2023finegrained, zhang2024cppo, wu2024private_rlhf, dai2024saferlhf, zhang2025pfppo}, leading to extended approaches such as GRPO~\citep{deepseek_math_2024}. For efficient training, DPO~\citep{rafailov2023DPO} has emerged as a primary choice because it significantly reduces computational overhead by eliminating the need for an explicit reward model. Within these methods, foundational probabilistic models for pairwise choice, such as the Bradley-Terry and Plackett-Luce models~\citep{bradley1952rank, plackett1975analysis}, remain standard for preference modeling 
and are typically applied at the level of complete responses.

\paragraph{Granularity of DPO.}
Several recent studies have proposed granularity-aware methods to better align with autoregressive generation.
For example,
SimPO employs a length-normalized loss function~\citep{meng2024SimPO},
TDPO imposes forward-KL constraints at each token position~\citep{zeng24tdpo}, 
TGDPO introduces token-level reward guidance~\citep{zhu25tgdpo}, and
cDPO~\citep{lin2025critical} identifies critical tokens to apply token-level weights.
Despite these advances, existing methods often fundamentally rely on the vanilla BT model at the level of complete responses.
In contrast, our formulation integrates the autoregressive assumption into DPO when applying the BT model.

Beyond these granularity-oriented approaches, \citet{rafailov2024from} provides a complementary analysis showing that standard DPO can be interpreted at the token level through a soft-Q-learning lens.
While this analysis effectively explains the implicit token-level structure of the original response-level BT formulation, it does not modify the BT model itself.
In contrast, our formulation incorporates the autoregressive assumption before applying the BT model by extending the energy domain to the prefix closure~$Y^{*}$, yielding a prefix-wise BT model and a distinct objective.

\section{Preliminary}
\label{seq:preliminary}

DPO~\cite{rafailov2023DPO} is a preference optimization algorithm that directly optimizes models using human preferences without relying on an explicit reward model.
To simplify the derivation and interpretation of its objective function, we reformulate DPO using two Boltzmann distributions.
We begin by defining the two reward-based energy functions, which we refer to as the DPO energies.

\textit{\paragraph{Definition 1 (DPO Energies).}
Given a reward function $r(x, y)$ and a reference model $\pi_{\mathrm{ref}} (y|x)$,
we define the two DPO energies, namely the likelihood energy $E_{1}$ and the posterior energy $E_{2}$:
\begin{align}
    E_{1}(x, y) &= - r(x, y), \label{eq:energy1} \\
    E_{2}(x, y) &= - \frac{1}{\beta} r(x, y) - \log \pi_{\mathrm{ref}} (y|x), \label{eq:energy2}
\end{align}
where
$x \in \mathcal{X}$ is an input from the input space $\mathcal{X}$, $y \in \mathcal{Y}$ is the model's output (response) in the output space $\mathcal{Y}$, and $\beta \in \mathbb{R}_{+}$ is a hyperparameter.
}

These energy functions derive the two corresponding Boltzmann distributions:
\begin{align}
    p_{1}(y^{w} \succ y^{l}|x) &= \frac{\exp(- E_{1}(x, y^{w}))}{\sum_{y \in Y} \exp(- E_{1}(x, y))},
\label{eq:blz1} \\
    p_{2}(y|x) &= \frac{\exp(- E_{2}(x, y))}{\sum_{y \in \mathcal{Y}} \exp(- E_{2}(x, y))},
\label{eq:blz2}
\end{align}
where $p_{1}$ is normalized over a set of preferred and dispreferred responses $Y=\{y^{w}, y^{l}\} \subset \mathcal{Y}$ following the BT model~\citep{bradley1952rank}, whereas $p_{2}$ is normalized over the entire
output space \(\mathcal{Y}\).
It is straightforward to see that $p_{2}$ maximizes the KL-constrained reward function~\citep{Peters2007reinforcement, peng2019advantage, Korbak2022reinforcement, go2023aligning, rafailov2023DPO}.
Then, DPO minimizes the negative log-likelihood loss $\mathcal{L}$ with respect to $p_{1}$:
\begin{align}
\hspace{-6pt}
&\mathcal{L}_{\text{DPO}}
\hspace{-1.9pt}
=
\hspace{-1.9pt}
- \mathbb{E}_{(x, Y) \sim \mathcal{D}}
\bigl[
\log p_{1}(y^{w}\!\succ\!y^{l}|x)
\bigr]
\hspace{-1.9pt} \nonumber \\
&=
\hspace{-1.9pt}
- \mathbb{E}_{(x, Y) \sim \mathcal{D}}
\hspace{-3pt}
\left[
\log \sigma
\hspace{-1pt}
\left(\beta \log \frac{p_{2}(y^{w}|x)}{\pi_{\mathrm{ref}}(y^{w}|x)}
\hspace{-2pt}-\hspace{-1pt}
\beta \log \frac{p_{2}(y^{l}|x)}{\pi_{\mathrm{ref}}(y^{l}|x)}\right)
\hspace{-2pt}
\right]
\hspace{-7pt}
\end{align}
where $\sigma$ is the sigmoid function and $\mathcal{D}$ is the preference dataset.
During training, $p_{2}$ is parameterized as $p_{2} = \pi_{\theta}$ with a parameter set $\theta$. This is the reparameterization trick of DPO.

In the recent paradigm of LLMs, models are typically assumed to be autoregressive, \textit{i.e.}, assuming that $\mathcal{Y}$ is a space of sequence of arbitrary length, both reference and learnable models satisfy the chain rule: $\pi(y|x) = \prod_{i=1}^{T} \pi(y_{i}|y_{<i}, x)$, where $T = \mu(y)$ is the length of $y$.\footnote{$\mu : \mathcal{Y} \to \mathbb{N}$ is the token length measure that assigns to each $y \in \mathcal{Y}$ the number of tokens it contains.}
From this, we have
\begin{align}
\label{eq:dpo_loss}
\mathcal{L}_{\text{DPO}}
=
- \mathbb{E}_{(x, Y) \sim \mathcal{D}}
\Biggl[
&\log \sigma\Bigl(
\sum_{i=1}^{T}
\Bigl(
\beta \log \frac{\pi_{\theta}(y^{w}_{i}| y^{w}_{<i}, x)}{\pi_{\mathrm{ref}}(y^{w}_{i}| y^{w}_{<i}, x)}
\nonumber \\
&-
\beta \log \frac{\pi_{\theta}(y^{l}_{i}| y^{l}_{<i}, x)}{\pi_{\mathrm{ref}}(y^{l}_{i}| y^{l}_{<i}, x)}
\Bigr)
\Bigr)
\Biggr],
\end{align}
where right-padding is applied so that both sequences have the same length following practical implementation of DPO using mini-batch optimization algorithms.

\section{Autoregressive Direct Preference Optimization}

From the derivation of DPO, we observe an important discrepancy: while the learnable model $\pi_{\theta}$ is autoregressive, the Boltzmann distribution $p_{2}$ defined in Eq.~(\ref{eq:blz2}) is not formulated as such.
This motivates our research question: \textit{Can we define energy functions that yield $p_{2}$ explicitly as an autoregressive distribution?}
Our ADPO formulation answers this question affirmatively.

The core idea behind ADPO lies in the introduction of the ADPO energies, which define energies and the corresponding implicit reward function over the prefix closure of the output space.

\textit{\paragraph{Definition 2 (ADPO Energies).}
Let $\mathcal{Y}^{*}$ be the prefix closure of $\mathcal{Y}$, i.e.,
\begin{align}
\mathcal{Y}^{*}
=
\bigcup_{y \in \mathcal{Y}}
\left\{
y_{\le i} = (y_{1}, y_{2}, \ldots, y_{i})
:\;
0 \le i \le T'
\right\},
\end{align}
where $T' = \mu'(y)$ is the length of $y$.\footnote{We will discuss later that this length measure can be independent of the token length measure $\mu$, but one can assume $\mu'=\mu$ for readability.}
Given a \uline{prefix-wise} reward function
$r^{*} : \mathcal{X} \times \mathcal{Y}^{*} \to \mathbb{R}$
and an \uline{autoregressive} reference model $\pi_{\mathrm{ref}}(y|x)$,
we define the two ADPO energies, namely the prefix likelihood energy $E_{1}^{*}$ and the prefix posterior energy $E_{2}^{*}$, as
\begin{align}
E_{1}^{*}(x, y_{\le i})
&=
- r^{*}(x, y_{\le i}),
\label{eq:a_energy1}
\\
E_{2}^{*}(x, y_{\le i})
&=
- \frac{1}{\beta} r^{*}(x, y_{\le i})
- \log \pi_{\mathrm{ref}}(y_{i} | y_{<i}, x).
\label{eq:a_energy2}
\end{align}
}

There are two main differences compared to Definition~1.
First, the reward function is defined over $\mathcal{X} \times \mathcal{Y}^{*}$ instead of $\mathcal{X} \times \mathcal{Y}$.
With this, we define the prefix-wise BT model by taking the product of BT-based preference distributions:
\begin{align}
p_{1}(y^{w} \succ y^{l} \mid x)
&=
\prod_{i=1}^{T'}
\frac{
\exp\!\left(- E_{1}^{*}(x, y^{w}_{\le i})\right)
}{
\sum_{y_{\le i} \in Y_{i}}
\exp\!\left(- E_{1}^{*}(x, y_{\le i})\right)
},
\label{eq:a_blz1}
\end{align}
where $Y_{i} = \{ y^{w}_{\le i}, y^{l}_{\le i} \} \subset \mathcal{Y}^{*}$ is the set of preferred and dispreferred prefixes up to length $i$.

Second, the reference model is assumed to be autoregressive in Eq.~(\ref{eq:a_energy2}).
With this assumption, the second distribution $p_{2}$ also becomes autoregressive prior to reparameterization:
\begin{align}
p_{2}(y|x)
&=
\prod_{i=1}^{T'} p_{2}(y_{i} \mid y_{<i}, x),
\\
p_{2}(y_{i} \mid y_{<i}, x)
&=
\frac{
\exp\!\left(- E_{2}^{*}(x, y_{\le i})\right)
}{
\sum_{y_{i} \in \mathcal{V}}
\exp\!\left(- E_{2}^{*}(x, y_{\le i})\right)
},
\label{eq:a_blz2}
\end{align}
where $\mathcal{V} = \{ y \in \mathcal{Y} : \mu'(y) = 1 \}$.
This alleviates the mismatch between $p_{2}$ and the learnable autoregressive model $\pi_{\theta}$.

\paragraph{ADPO Objective.}
Analogous to DPO, ADPO maximizes the log-probability of $p_{1}(y^{w} \succ y^{l} \mid x)$ during training.
We define the loss function as
\begin{align}
\mathcal{L}_{\text{ADPO}}
=
- \mathbb{E}_{(x, Y) \sim \mathcal{D}}
\Bigl[
\log p_{1}(y^{w} \succ y^{l} \mid x)
\Bigr].
\end{align}
Through reparameterization, the loss can be written as
\begin{align}
\label{eq:adpo_loss}
\mathcal{L}_{\text{ADPO}}
=
- \mathbb{E}_{(x, Y) \sim \mathcal{D}}
&\Biggl[
\sum_{i=1}^{T'}
\log \sigma \Bigl(
\beta
\log
\frac{
\pi_{\theta}(y^{w}_{i} \mid y^{w}_{<i}, x)
}{
\pi_{\mathrm{ref}}(y^{w}_{i} \mid y^{w}_{<i}, x)
}
\nonumber \\
&-
\beta
\log
\frac{
\pi_{\theta}(y^{l}_{i} \mid y^{l}_{<i}, x)
}{
\pi_{\mathrm{ref}}(y^{l}_{i} \mid y^{l}_{<i}, x)
}
\Bigr)
\Biggr].
\end{align}
A proof is provided in Appendix~\ref{app:adop_loss}.
Notably, ADPO shifts the summation in the DPO loss (Eq.~(\ref{eq:dpo_loss})) outside the log-sigmoid function.
In terms of KL-constrained reward maximization, the optimal solution is preserved (Appendix~\ref{app:klmax}).

\section{Theoretical Analysis}

ADPO does not violate the theoretical foundations of DPO.
Rather, the theory is generalized by bridging additive reward functions and autoregressive models.
Ultimately, we prove that, under mild assumptions, any reward function can be reparameterized by an autoregressive model (\textbf{Theorem~1}).
This better aligns the DPO theory with the paradigm of autoregressive LLMs.

\subsection{Reparameterization Completeness}

We first show how prefix-wise reward functions can be reparameterized by autoregressive models.

\textit{\textbf{Proposition 1 (Prefix-wise Reparameterization Completeness).}
Let $[r^{*}]$ denote the reward-shift equivalence class of a prefix-wise reward function
$r^{*} : \mathcal{X} \times \mathcal{Y}^{*} \to \mathbb{R}$, defined as
\begin{align}
[r^{*}]
=
\Bigl\{
r'
\;\Big|\;
&\exists f \;\;
\forall (x, y_{\le i}), \nonumber \\
& r'(x, y_{\le i})
=
r^{*}(x, y_{\le i})
+
f(x, y_{<i})
\Bigr\}.
\end{align}
Given an autoregressive reference model $\pi_{\mathrm{ref}}(y_{i}\mid y_{<i}, x) > 0$ and a hyperparameter $\beta > 0$,
for any prefix-wise reward function $r^{*}$, there exists a unique representative
$r^{*}_{\circ} \in [r^{*}]$ such that for all $x \in \mathcal{X}$ and prefixes $y_{\le i} \in \mathcal{Y}^{*}$,
\begin{align}
\label{eq:representative}
r^{*}_{\circ}(x, y_{\le i})
\equiv
\beta \log
\frac{
\pi(y_{i} \mid y_{<i}, x)
}{
\pi_{\mathrm{ref}}(y_{i} \mid y_{<i}, x)
},
\end{align}
for some \uline{autoregressive} model $\pi$.}

A proof is provided in Appendix~\ref{app:completeness}.
To bridge vanilla reward functions $r(x, y)$ and prefix-wise reward functions $r^{*}(x, y_{\le i})$, we define the additive decomposition of $r$.
Lemma~1 immediately follows.

\textit{\textbf{Definition 3 (Additive Decomposition).}
Let $r:\mathcal{X} \times \mathcal{Y} \to \mathbb{R}$ be a reward function.
We say that a prefix-wise reward function
$r^{*}:\mathcal{X} \times \mathcal{Y}^{*} \to \mathbb{R}$
is an additive decomposition of $r$ if and only if
\begin{align}
r(x, y)
=
\sum_{i=1}^{T'} r^{*}(x, y_{\le i})
\quad
\text{for any } x \in \mathcal{X},\, y \in \mathcal{Y}.
\end{align}
}

\textit{\textbf{Lemma 1.}
Every reward function has an additive decomposition.
}

Then, from Proposition~1 and Lemma~1, it follows that Theorem~1 holds.

\textit{\textbf{Theorem 1.}
All reward classes consistent with the \uline{prefix-wise} Bradley--Terry models can be represented with the reparameterization
\begin{align}
r(x, y)
=
\beta \log
\frac{\pi(y \mid x)}{\pi_{\mathrm{ref}}(y \mid x)}
\end{align}
for some \uline{autoregressive} model $\pi(y\mid x)$ and a given \uline{autoregressive} reference model $\pi_{\mathrm{ref}}(y\mid x)$ such that $\pi(y_{i} \mid y_{<i}, x) > 0$.
}

\begin{proof}
For any reward function $r$, an additive decomposition $r^{*}$ exists by Lemma~1.
By Proposition~1, there exists an autoregressive model $\pi$ such that
\begin{align}
\label{eq:representative_}
r(x, y)
&=
\sum_{i=1}^{T'} r^{*}(x, y_{\le i})
\equiv
\sum_{i=1}^{T'} r^{*}_{\circ}(x, y_{\le i})
\nonumber \\
&=
\sum_{i=1}^{T'}
\beta \log
\frac{
\pi(y_{i} \mid y_{<i}, x)
}{
\pi_{\mathrm{ref}}(y_{i} \mid y_{<i}, x)
}
=
\beta \log
\frac{
\pi(y \mid x)
}{
\pi_{\mathrm{ref}}(y \mid x)
},
\end{align}
where $\equiv$ denotes equality within the equivalence class.
\end{proof}

In contrast to the DPO theory~\citep{rafailov2023DPO}, Theorem~1 explicitly demonstrates reparameterizability using an autoregressive model.

\subsection{Implicit length measure in DPO}

The following corollary further bridges DPO and ADPO, providing deeper insights into DPO.

\textit{\textbf{Corollary 1.}
When the length $\mu'(y)$ is one for all $y \in \mathcal{Y}$,
ADPO reduces to DPO that relies on the traditional (non-prefixwise) Bradley--Terry model.
}

This result can be unintuitive.
A straightforward interpretation of ``$\mu'(y)=1$ for all $y$'' is that the output space $\mathcal{Y}$ is restricted to a single-token space, implying that ADPO coincides with DPO in this special case.
While this interpretation is correct, it does not fully capture the true meaning of the corollary.
Instead, we emphasize that the output space $\mathcal{Y}$ remains a sequence space.
Then, Corollary~1 implies that the original DPO formulation has an implicit length measure
$\mu' : \mathcal{Y} \to \mathbb{N}$ that assigns a length of one, i.e., $\mu'(y)=1$, to every sequence $y \in \mathcal{Y}$.

This measure may have limited the exploration of research ideas for DPO-based formulations in prior work, as it predisposed most existing methods to place the summation operation inside the log-sigmoid function, as in Eq.~(\ref{eq:dpo_loss}).
We further discuss its deeper roots.

\begin{figure*}
\centering
\includegraphics[width=\linewidth]{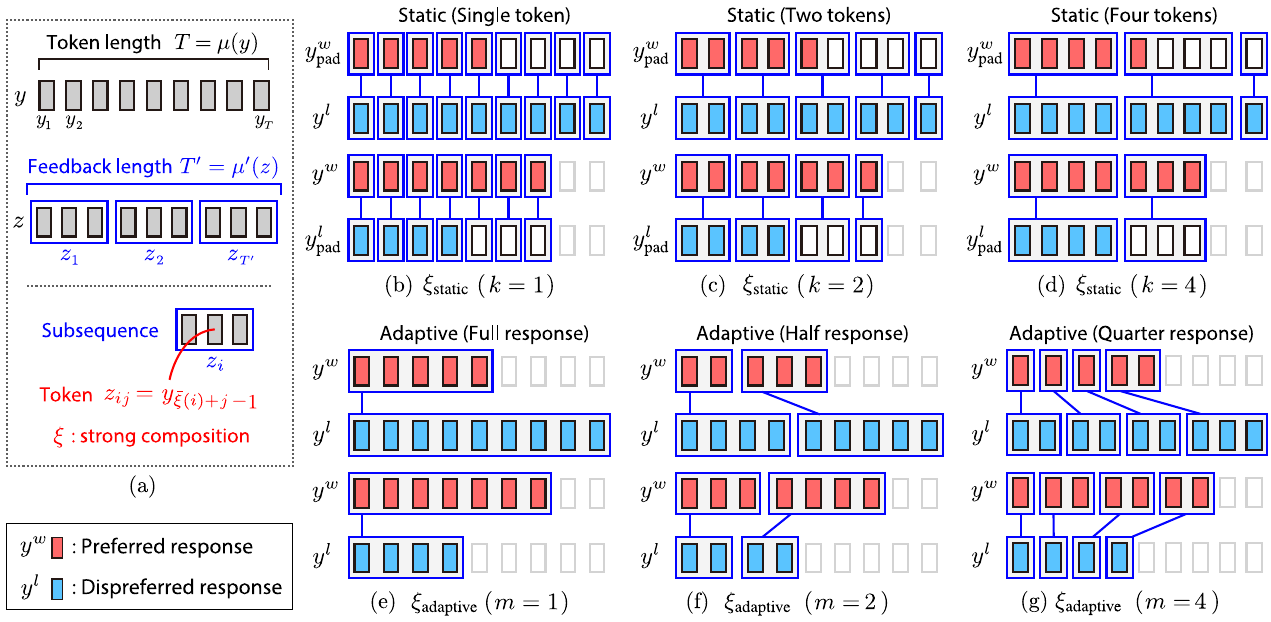}
\caption{
Static and adaptive families of ADPO.
(a) Token and feedback length measures.
Each subsequence $z_{i}$ is defined by a strong composition $\xi$.
(b--d) Static families with a fixed window size $k$.
(e--g) Adaptive families with a fixed number of subsequences.
(e) corresponds to DPO.
Blue rectangles indicate regions where summation is applied inside the log-sigmoid function.
}
\label{fig:gra}
\end{figure*}

\subsection{Two Distinct Length Measures}

\paragraph{What is $\bm{\mu'}$?}
Although the implicit length measure $\mu'$ may appear counterintuitive, it is natural to interpret it as arising from RLHF.
Concretely, $\mu'$ measures the length of sequences $y$ in an evaluation scenario by first mapping $y$ into a one-dimensional space using an evaluation metric $\nu : \mathcal{Y} \to \mathbb{R}$, and then defining
\begin{align}
\mu'(y)
=
\mathrm{length}(\nu(y))
=
\mathrm{dim}(\mathbb{R})
=
1.
\end{align}
This interpretation is particularly natural when human feedback is performed on complete responses.
Therefore, we refer to $\mu'$ as the \textit{feedback length measure}.

\paragraph{Two Length Measures are Independent.}
The discussion thus far suggests two algorithms.
The first is DPO, corresponding to the case $\mu' \equiv 1$, which evaluates the complete response as a whole; thus, we denote its granularity as ``Full Response''.
The second is token-level ADPO, corresponding to $\mu'=\mu$, where $\mu$ is the token length measure induced by LLMs.
This evaluates each token individually, and its granularity is indicated as ``Single Token''.

However, importantly, ADPO itself does not impose the constraint $\mu'=\mu$.
Although the two measures should be related in practice, they can be theoretically independent because their origins are independent: one arises from LLM tokenization and the other from the prefix closure.
This leads to granularity families of ADPO, which we discuss in the next subsection.

\subsection{Granularity Families}

ADPO allows intermediate granularity levels with
$1 \le \mu'(y) \le \mu(y)$ in practice.
Specifically, by decomposing each sequence $y$ into subsequences
$\{ z_{i} \}_{i=1}^{T'}$ such that each prefix
$z_{\le i} = (z_{1}, z_{2}, \ldots, z_{i})$
constitutes a unit eligible for implicit feedback,
the ADPO loss can be formulated with two cumulative log-ratios $S_{\theta}^{w}$ and $S_{\theta}^{l}$:

\begin{align}
S_{\theta}^{w}(i)
&\triangleq
\sum_{j=1}^{T_{i}^{w}}
\log
\frac{
\pi_{\theta}(z^{w}_{i,j} \mid z^{w}_{<i,<j}, x)
}{
\pi_{\mathrm{ref}}(z^{w}_{i,j} \mid z^{w}_{<i,<j}, x)
},\label{eq:log_ratio_w}
\end{align}
\begin{align}
S_{\theta}^{l}(i)
&\triangleq
\sum_{j=1}^{T_{i}^{l}}
\log
\frac{
\pi_{\theta}(z^{l}_{i,j} \mid z^{l}_{<i,<j}, x)
}{
\pi_{\mathrm{ref}}(z^{l}_{i,j} \mid z^{l}_{<i,<j}, x)
}. \label{eq:log_ratio_l} \\
\mathcal{L}_{\text{ADPO}}
&=
- \mathbb{E}_{(x, Y) \sim \mathcal{D}}
\Biggl[
\sum_{i=1}^{T'}
\log \sigma \Bigl(
\beta S_{\theta}^{w}(i)
-
\beta S_{\theta}^{l}(i)
\Bigr)
\Biggr].\label{eq:loss_adpo_final}
\end{align}

where $z = (z_{1}, z_{2}, \ldots, z_{T'})$ is a decomposition of $y$,
$z_{i} = (z_{i,1}, z_{i,2}, \ldots, z_{i,T_{i}})$ is a subsequence,\footnote{
Each prefix is denoted by
$z_{<i,<j}
=
(z_{1}, \ldots, z_{i-1},
(z_{i,1}, \ldots, z_{i,j-1}))$.
}
$T'$ is the feedback length,
and $T_{i} = \mu(z_{i})$ is the token length of $z_{i}$.

More formally, the mapping from $y$ to $z$ is specified by a strong composition $\xi$.
Given a token sequence
$y = (y_{1}, \ldots, y_{T}) \in \mathcal{Y}$,
each subsequence is defined as
\begin{align}
z_{i}
=
\bigl(
y_{\bar{\xi}(i)},
y_{\bar{\xi}(i)+1},
\ldots,
y_{\bar{\xi}(i+1)-1}
\bigr),
\end{align}
with
\begin{align}
\bar{\xi}(1) &= 1, \\
\bar{\xi}(i+1) &= \bar{\xi}(i) + \xi(i),
\end{align}
where $\xi \in \mathcal{C}_{T'}(T)$ and
\begin{align}
\mathcal{C}_{m}(n)
=
\Bigl\{
\xi : [m] \to \mathbb{N}_{+}
\;\Big|\;
\sum_{j=1}^{m} \xi(j) = n
\Bigr\}.
\end{align}
is the set of strong compositions of $n \in \mathbb{N}_{+}$ into $m \in \mathbb{N}_{+}$ parts.\footnote{
$[m] = \{1, 2, \ldots, m\}$.
}
Each token is then given by
$z_{i,j} = y_{\bar{\xi}(i) + j - 1}$,
as illustrated in Fig.~\ref{fig:gra}(a).
The choice of $\xi$ determines the learning behavior of ADPO.
We define two practical families below.

\paragraph{Static Family.}
This family fixes a window size $k \in \mathbb{N}_{+}$ and determines the feedback length based on the token length.
Because preferred and dispreferred responses may have different lengths,
we apply EOS padding to equalize them, obtaining
$y_{\text{pad}}^{w}$ and $y_{\text{pad}}^{l}$.
Each sequence is then decomposed into $\lceil T/k \rceil$ subsequences,
where
$T = \mu(y_{\text{pad}}^{w}) = \mu(y_{\text{pad}}^{l})$.
The strong composition is defined as
\begin{align}
\xi_{\text{static}}(i)
=
\min \bigl(
k,\,
\mu(y_{\text{pad}}^{w}) - k(i-1)
\bigr).
\end{align}
As shown in Fig.~\ref{fig:gra}(b--d),
$k=1$ corresponds to token-level granularity,
while larger $k$ yields coarser feedback.

\paragraph{Adaptive Family.}
This family decomposes each sequence into
$m \in \mathbb{N}_{+}$ subsequences as uniformly as possible
and fixes the feedback length as $\mu'(y)=m$ for all $y \in \mathcal{Y}$.
The strong composition is defined as
\begin{align}
\xi_{\text{adaptive}}(i)
=
\left\lfloor \frac{i\,\mu(y)}{m} \right\rfloor
-
\left\lfloor \frac{(i-1)\,\mu(y)}{m} \right\rfloor.
\end{align}
When $m=1$, this formulation reduces to DPO, as shown in Fig.~\ref{fig:gra}(e).
For $m>1$, preferred and dispreferred responses are partitioned uniformly
(Fig.~\ref{fig:gra}(f,g)),
naturally extending DPO to finer granularity.

\begin{table*}[t]
\setlength{\tabcolsep}{3.2pt}
\small
\centering
\caption{Experimental results on two mathematical reasoning benchmarks.}
\begin{tabular}{lcccccccc}
\toprule
\multirow{2}{*}{\textbf{Method}} 
& \multicolumn{2}{c}{\textbf{Llama-3-8B}} 
& \multicolumn{2}{c}{\textbf{Gemma-3-12B}} 
& \multicolumn{2}{c}{\textbf{Qwen-3-8B}} 
& \multicolumn{2}{c}{\textbf{DeepSeek-Math-7B}} \\[0.1em]
\cmidrule(lr){2-3} \cmidrule(lr){4-5} \cmidrule(lr){6-7} \cmidrule(lr){8-9}
& GSM8K & MATH 
& GSM8K & MATH 
& GSM8K & MATH 
& GSM8K & MATH \\
\midrule
DPO~\citep{rafailov2023DPO}
& 64.37 & 18.00 & 77.03 & 39.80 & 86.96 & 53.80 & 67.78 & 32.00
\\[0.1em]
\rowcolor{gray!12}
ADPO (Ours) &
\textbf{68.08} & \textbf{21.00} & \textbf{78.32} & \textbf{41.20} & \textbf{88.10} & \textbf{55.40} & \textbf{69.98} & \textbf{33.40}\\
\midrule
cDPO~\citep{lin2025critical} &
67.90 & 16.80 &
77.18 & 38.60 &
90.98 & 56.80 & 
72.90 & 33.40 \\[0.1em] 
\rowcolor{gray!12}
cADPO (Ours)
& \textbf{68.76} & \textbf{20.20} &
\textbf{78.85} & \textbf{40.40} & \textbf{91.74} & \textbf{57.20} & 
\textbf{73.54} & \textbf{35.40} \\ 
\bottomrule
\end{tabular}
\label{tab:math}
\end{table*}
\definecolor{lightgray}{gray}{0.8}
\begin{table*}
\setlength{\tabcolsep}{1pt}
\small
\centering
\caption{Granularity families of ADPO.
Best results within the static and adaptive families are underlined, and the overall best results are highlighted in bold.
}
\label{tab:granularity_analysis}
\begin{tabular}{lcccp{0.8cm}p{0.9cm}p{0.8cm}p{1.0cm}p{0.8cm}p{0.9cm}p{0.8cm}p{0.9cm}}
\toprule
\multirow{2}{*}{\textbf{Method}} & \multirow{2}{*}{$\mu'(y)$} & \multirow{2}{*}{\textbf{Composition}} & \multirow{2}{*}{\textbf{Granularity}} &
\multicolumn{2}{p{1.7cm}}{\textbf{Llama-3-8B}} & \multicolumn{2}{p{1.7cm}}{\scalebox{0.95}[1.0]{\textbf{\hspace{-3.5pt}Gemma-3-12B}}} & \multicolumn{2}{p{1.7cm}}{\textbf{Qwen-3-8B}} & \multicolumn{2}{p{1.8cm}}{\textbf{DS-Math-7B}}
\\
&&&& \scalebox{0.9}{GSM} & \scalebox{0.9}{MATH} & \scalebox{0.9}{GSM} & \scalebox{0.9}{MATH} & \scalebox{0.9}{GSM} & \scalebox{0.9}{MATH} & \scalebox{0.9}{GSM} & \scalebox{0.9}{MATH}\\
\midrule
ADPO & $\mu(y)/8$ & $\xi_{\text{static}}~(k\!=\!8)$ & Eight tokens &
64.97 & 18.20 & 76.88 & 40.00 & 87.79 & 54.20 & 68.39 & 32.00\\
ADPO & $\mu(y)/4$ & $\xi_{\text{static}}~(k\!=\!4)$ & Four tokens & 66.26 & 17.40 & 77.41 & 40.80 & \underline{88.48} & 54.20 & 68.23 & 32.00\\
ADPO & $\mu(y)/2$ & $\xi_{\text{static}}~(k\!=\!2)$ & Two tokens & \underline{66.57} & \underline{18.60} & 77.63 & \underline{\textbf{41.20}} & 88.25 & 54.20 & 70.05 & 33.60 \\
ADPO & $\mu(y)$ & $\xi_{\text{static}}~(k\!=\!1)$ & Single token & 66.41 & 18.00 & \underline{78.09} & 39.80 & \underline{88.48} & \underline{54.40} & \underline{\textbf{70.36}} & \underline{\textbf{34.80}}\\
\arrayrulecolor{lightgray}
\midrule
\arrayrulecolor{black}
DPO & 1 & $\xi_{\text{adaptive}} (m\!=\!1)$ & Full response & 64.37 & 18.00 & 77.03 & 39.80 & 86.96 & 53.80 & 67.78 & 32.00\\
ADPO & 2 & $\xi_{\text{adaptive}} (m\!=\!2)$ & Half response & 64.29 & 19.60 & 76.88 & 40.20 & 87.95 & 53.80 & 68.61 & 33.40 \\
ADPO & 16 & $\xi_{\text{adaptive}} (m\!=\!16)$ & 1/16 response & 64.52 & 18.00 & 77.10 & 40.80 & 87.26 & 52.40 & 68.84 & 33.40\\
ADPO & 128 & $\xi_{\text{adaptive}} (m\!=\!128)$ & 1/128 response & 66.72 & 18.20 & 77.63 & 41.00 & \underline{\textbf{89.16}} & 54.60 & 69.83 & \underline{33.60}\\
ADPO & 256 & $\xi_{\text{adaptive}} (m\!=\!256)$ & 1/256 response & \underline{\textbf{68.08}} & \underline{\textbf{21.00}}
& \underline{\textbf{78.32}} & \underline{\textbf{41.20}} & 88.10 & \underline{\textbf{55.40}} & 69.98 & 33.40\\
ADPO & 512 & $\xi_{\text{adaptive}} (m\!=\!512)$ & 1/512 response & 67.32 & 19.00 & 77.94 & 40.00 & 88.02 & 54.00 & \underline{\textbf{70.36}} & 34.40\\
\bottomrule
\end{tabular}
\end{table*}

\section{Experiments}

\subsection{Experimental Setup}

\paragraph{Datasets and Metrics.}
We use two representative math reasoning datasets: GSM8K~\citep{cobbe2021gsm8k} and MATH500~\citep{hendrycks2021math}.
GSM8K is a dataset of grade school math word problems that requires multi-step reasoning to arrive at the correct numerical answer.
MATH comprises challenging competition mathematics problems spanning various topics including algebra, geometry, probability, and number theory.
All models are fine-tuned on the GSM8K training set.
We report accuracy
using the 8-shot evaluation protocol for GSM8K~\citep{wei2022CoT} and the 4-shot evaluation protocol for MATH500~\citep{lewkowycz2022solving}.

\paragraph{Baselines.} 
We include two baselines:
DPO~\citep{rafailov2023DPO} and cDPO~\citep{lin2025critical}.
cDPO is one of the latest preference optimization methods, which leverages token-level contrastive estimation to identify and penalize critical tokens in incorrect reasoning trajectories.

\paragraph{ADPO Variants.}
We apply our approach to both DPO and cDPO, and refer to the resulting variants as ADPO and cADPO, respectively.
ADPO uses the loss in Eq.~(\ref{eq:loss_adpo_final}) and cADPO extends it to utilize the critical tokens.
Specifically, we define
\begin{align}
\tilde{S}_{\theta}^{l}(i)
\triangleq
\sum_{j=1}^{T_{i}^{l}}
\bar{s}_{j}
\log
\frac{
\pi_{\theta}(z^{l}_{i,j} \mid z^{l}_{<i,<j}, x)
}{
\pi_{\mathrm{ref}}(z^{l}_{i,j} \mid z^{l}_{<i,<j}, x)
},
\label{eq:log_ratio_l_weighted}
\end{align}
where $\bar{s}_{j} = 1 - s_{j}$ is a token-level weight computed from the reward score $s_{j}$ obtained via contrastive estimation~\citep{lin2025critical}.
Using $S_{\theta}^{w}(i)$ defined in Eq.~(\ref{eq:log_ratio_w}) and the weighted log-ratio $\tilde{S}_{\theta}^{l}(i)$, the cADPO loss is given by
\begin{align}
\label{eq:loss_cadpo_final}
\mathcal{L}_{\text{cADPO}}
=
- \mathbb{E}_{(x, Y) \sim \mathcal{D}}
\Biggl[
\sum_{i=1}^{T'}
\log \sigma \Bigl(
\beta S_{\theta}^{w}(i)
-
\beta \tilde{S}_{\theta}^{l}(i)
\Bigr)
\Biggr].
\end{align}
We use ADPO and cADPO in the adaptive family with $m=256$ as our primary methods.

\paragraph{Models.} We conduct experiments with four diverse LLMs: Llama3-8B-Base~\citep{llama3_herd_2024},
Qwen-3-8B-Base~\citep{qwen3_techreport_2025}, Gemma-3-12B-PT~\citep{gemmateam2025gemma3} and DeepSeek-Math-7B~\citep{deepseek_math_2024}.

\paragraph{Implementation Details.}
Each model is trained with LoRA using the AdamW optimizer for 3 epochs.
Since cDPO requires a larger learning rate, we set it to $2.0 \times 10^{-5}$ for DPO and ADPO, and $4.0 \times 10^{-5}$ for cDPO and cADPO.

\subsection{Experimental Results}

\paragraph{Main Results.}
Table~\ref{tab:math} summarizes the experimental results.
We observe that our approach consistently outperforms both the DPO and cDPO baselines across all four LLMs, achieving the best results with cADPO.
The highest individual result was obtained by Qwen-3-8B using cADPO, reaching 91.74\% on GSM8K.
These results demonstrate the practical effectiveness of our approach.

\paragraph{Granularity Families.}
To analyze how varying granularity affects ADPO performance, we systematically explore both static and adaptive families across a range of hyperparameter values in Table~\ref{tab:granularity_analysis}.
For the static family, we observe that models with $k \leq 4$ consistently outperform DPO, indicating the benefit of finer granularity.
With DeepSeek-Math-7B, the finest granularity (token-level, $k\!=\!1$)
achieves the best performance.
However, there remains room for improvement, as the static family requires padding when computing the loss, which potentially constrains its performance.

For the adaptive family, which is equivalent to DPO when $m=1$, we observe performance degradation across several LLMs when $m$ is small (\textit{e.g.}, $m=2$).
This may occur because dividing responses into a small number of segments can result in unnatural or arbitrary splits.
This limitation is addressed by increasing the number of segments.
Specifically, when $m \geq 128$, ADPO consistently outperforms DPO in all cases, again indicating the benefit of finer granularity.

When comparing the best-performing models from the static and adaptive families, the adaptive family tends to achieve superior performance.
As the adaptive family naturally extends DPO without requiring padding, it effectively addresses the limitations associated with the static family.
Overall, these findings validate our theoretical insight that defining the token-length measure and feedback-length measure through strong compositions is effective.

\begin{figure*}[t]
\centering
\includegraphics[width=\linewidth]{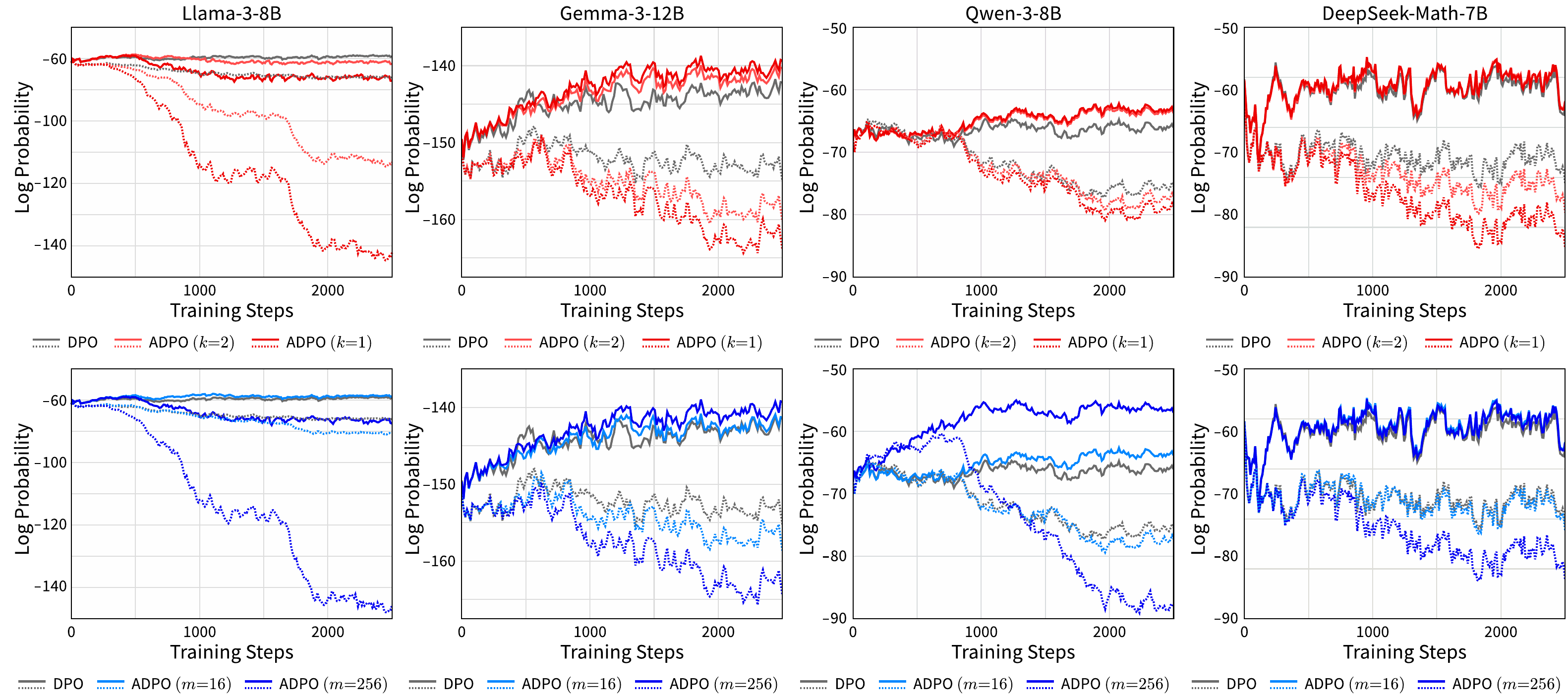}
\caption{
Comparison of training dynamics between DPO and ADPO.
The evolution of log probabilities for preferred sequences (solid line) and dispreferred sequences (dashed line) during training is shown. Top row: static family with $k=1, 2$. Bottom row: adaptive family with $m=16, 256$.
}
\label{fig:training}
\end{figure*}

\paragraph{Learning Behavior.}
Figure~\ref{fig:training} analyzes the training dynamics of ADPO compared to DPO. Specifically, it shows the evolution of log probabilities for preferred (chosen) and dispreferred (rejected) sequences across training steps for each LLM.
We observe that ADPO effectively amplifies the distinction between preferred and dispreferred responses by consistently elevating the probabilities of preferred responses while suppressing those of dispreferred responses.
This tendency becomes more pronounced as granularity increases; that is, as $k$ decreases and $m$ increases.
These observations underscore the effectiveness of incorporating the autoregressive assumption into the BT model, enabling ADPO variants to more effectively guide models toward preferred outputs across training.

\begin{table*}[t]
\centering
\begin{minipage}{0.66\textwidth}
\setlength{\tabcolsep}{2.5pt}
\small

\iftwollms
\else
\caption{Experimental results on
AlpacaEval~2, Arena-Hard, and MT-Bench.
Following the official evaluation protocol of each benchmark, length-controlled win rate (LC), win rate (WR), and benchmark scores are reported.
}
\label{tab:conversation}
\centering
\begin{tabular}{lcccc}
\toprule
\multirow{3}{*}{\raisebox{-8pt}{\textbf{Method}}} 
& \multicolumn{4}{c}{\textbf{Llama3-8B}} \\
\cmidrule(lr){2-5}
& \multicolumn{2}{c}{\textbf{AlpacaEval~2}} & \textbf{Arena-Hard} & \textbf{MT-Bench} \\
\cmidrule(lr){2-3} \cmidrule(lr){4-4} \cmidrule(lr){5-5} 
& LC & WR & WR & Score \\
\midrule
SFT & 26.0 & 25.3 & 22.3  & 6.9 \\
DPO
\cite{rafailov2023DPO} & 40.3 & 37.9 & 32.6 & 7.0\\
SimPO
\cite{meng2024SimPO}
& 44.7 & 40.5 & 33.8 & 7.0\\
\rowcolor{gray!12}
ADPO (Ours) & \textbf{45.8} & \textbf{41.1} & \textbf{34.4} & \textbf{7.1}\\
\bottomrule
\end{tabular}
\fi

\end{minipage}\hfill
\begin{minipage}{0.32\textwidth}
\setlength{\tabcolsep}{4.5pt}
\small
\centering
\caption{Granularity families (AlpacaEval 2, Llama3-8B).}
\label{tab:gra_alpaca2}
\begin{tabular}{lcc}
\toprule
\textbf{Composition} & \textbf{LC} & \textbf{WR} \\[-0.2em]
\midrule
$\xi_{\text{static}}~(k\!=\!4)$ & 44.9 & 40.9 \\
$\xi_{\text{static}}~(k\!=\!2)$ & 45.1 & 40.8 \\
$\xi_{\text{static}}~(k\!=\!1)$ & \textbf{45.8} & 40.8 \\
\arrayrulecolor{lightgray}
\midrule
\arrayrulecolor{black}
$\xi_{\text{adaptive}}~(m\!=\!1)$ & 40.3 & 37.9\\
$\xi_{\text{adaptive}}~(m\!=\!2)$ & 41.0 & 40.2 \\
$\xi_{\text{adaptive}}~(m\!=\!16)$ & 44.1 & 40.2 \\
$\xi_{\text{adaptive}}~(m\!=\!128)$ & \textbf{45.8} & \textbf{41.1} \\
$\xi_{\text{adaptive}}~(m\!=\!256)$ & 44.6 & 40.5 \\
$\xi_{\text{adaptive}}~(m\!=\!512)$ & 44.7 & 40.3 \\
\bottomrule
\end{tabular}
\end{minipage}
\end{table*}

\paragraph{Conversation Tasks.}
To further validate the generalizability of our approach, we conduct experiments on conversation tasks.
We follow the setting of SimPO~\citep{meng2024SimPO}.
Specifically, we adopt three widely recognized benchmarks: AlpacaEval~2~\citep{dubois2024AlpacaEval}, Arena-Hard~\citep{li2025arenahard}, and MT-Bench~\citep{zheng2023MTBench}.
For the training data, we generate multiple candidate responses from the base model by prompting it with inputs from UltraFeedback~\citep{cui2024ultrafeedbackboostinglanguagemodels}, and construct chosen–rejected pairs based on PairRM annotations~\citep{dongfu2023parirm}.
We fine-tune Llama-3-8B-Instruct on this dataset and compare SFT, DPO~\citep{rafailov2023DPO}, SimPO, and ADPO.
Note that cDPO cannot perform conversation tasks, as it requires explicit True/False labels for each response, which are not available in UltraFeedback.
Table~\ref{tab:conversation} summarizes the results.
Consistently, ADPO achieves superior performance against DPO across all three benchmarks and performs comparably to or even better than SimPO, underscoring its enhanced capability in generating more preferred responses.

In Table~\ref{tab:gra_alpaca2}, we also find that granularity exhibits a similar trend as in mathematical reasoning tasks: finer-grained ADPO variants consistently outperform coarser granularity configurations in both static and adaptive families.
Collectively, these results demonstrate that ADPO not only improves mathematical reasoning but also effectively generalizes across diverse conversational tasks.
Additional experimental results are in Appendix~\ref{app:exp}.

\section{Conclusion}
We introduced ADPO, a novel approach for direct preference optimization that integrates the autoregressive assumption when applying the Bradley-Terry model.
By forming the prefix closure of the output space, we naturally derived the objective function of ADPO that shifts the summation operation outside the log-sigmoid function.
Moreover, we provided a theoretical analysis of ADPO, which yields deeper insights into the original DPO framework, identifying two distinct length measures.
Our experimental evaluations extensively validated ADPO across five benchmarks.

\paragraph{Limitations and Future Research Directions.}
While our study offered novel theoretical and practical perspectives on DPO, it still relies on the original KL-constrained reward maximization problem.
Extending beyond this constraint, future research could investigate more general divergence metrics as alternatives to KL divergence, potentially offering greater flexibility and robustness.
Moreover, our formulation of the prefix posterior energy in Eq.~(\ref{eq:a_energy2}) is specifically motivated by the autoregressive assumption characteristic of LLMs.
It would be valuable to explore alternative definitions of energy that are appropriate for other classes of generative models, particularly those that do not adhere strictly to autoregressive architectures.
Such investigations could further expand the applicability and effectiveness of DPO-based optimization methods across diverse model types.
Finally, ADPO introduces additional prefix-wise computations. Although we empirically observe that the resulting overhead is small (see Appendix~\ref{app:exp}), developing more efficient implementations remains an interesting direction.

\section*{Impact Statement}
This paper presents work whose goal is to advance the field of Machine
Learning. There are many potential societal consequences of our work, none
which we feel must be specifically highlighted here.

\section*{Acknowledgements}
This work was supported by JSPS KAKENHI Grant Numbers 25K03135 and 25H01137. These research results were obtained from the commissioned research (No.22501) by National Institute of Information and Communications Technology (NICT), Japan. This study was carried out using the TSUBAME4.0 supercomputer at Institute of Science Tokyo.


\bibliography{example_paper}
\bibliographystyle{icml2026}

\newpage
\appendix
\onecolumn


\section{Derivation of ADPO loss}
\label{app:adop_loss}

We provide a proof of Eq.~(\ref{eq:adpo_loss}).
We first provide detailed definition of ADPO loss and then show the Proposition A.1.

\textit{\textbf{Definition A.1.}
We define ADPO loss as
\begin{align}
\mathcal{L}_{\text{ADPO}}
=
- \mathbb{E}_{(x, Y) \sim \mathcal{D}}
\bigl[
\log p_{1}(y^{w} \succ y^{l}|x)
\bigr].
\end{align}
with ADPO energies given by
\begin{align}
E_{1}^{*}(x, y_{\leq i}) &= - r^{\circ}(x, y_{\leq i}),
\label{eq:a_energy1_app}\\
E_{2}^{*}(x, y_{\leq i}) &= - \frac{1}{\beta} r^{\circ}(x, y_{\leq i}) - \log \pi_{\mathrm{ref}} (y_{i}|y_{<i}, x),
\label{eq:a_energy2_app}
\end{align}
where $r^{\circ} \in [r^{*}]$ is a reward equivalent to $r^{*}$ given by
\begin{align}
r^{\circ} (x, y_{\leq i})
&= r^{*}(x, y_{\leq i})
- f(x, y_{< i}),\\
f(x, y_{< i})
&=
\beta \log \sum_{y_{i} \in \mathcal{V}}
\pi_{\text{ref}} (y_{i}|y_{<i}, x)
\exp\left(\frac{1}{\beta} r^{*}(x, y_{\leq i}) \right).
\end{align}
}

\textit{\textbf{Proposition A.1.}
Given an autoregressive reference model $\pi_{\mathrm{ref}}$, an autoregressive learnable model $\pi_{\theta}$, a dataset $\mathcal{D}$, and a constant $\beta > 0$, the ADPO loss is given by
\begin{align}
\mathcal{L}_{\text{ADPO}}
&=
- \mathbb{E}_{(x, Y) \sim \mathcal{D}}
\left[
\sum_{i=1}^{T'}
\log \sigma\left( \beta \log \frac{\pi_{\theta}(y^{w}_{i}|y^{w}_{<i}, x)}{\pi_{\mathrm{ref}} (y^{w}_{i}|y^{w}_{<i}, x)}
- \beta \log \frac{\pi_{\theta}(y^{l}_{i}|y^{l}_{<i}, x)}{\pi_{\mathrm{ref}} (y^{l}_{i}|y^{l}_{<i}, x)} \right)
\right], \tag{\ref{eq:adpo_loss}}
\end{align}
where $\sigma$ is the sigmoid function.
}
\begin{proof}
From the definition of ADPO loss, we have
\begin{align}
\mathcal{L}_{\text{ADPO}}
&=
- \mathbb{E}_{(x, Y) \sim \mathcal{D}}
\bigl[
\log p_{1}(y^{w} \succ y^{l}|x)
\bigr]\\
&=
- \mathbb{E}_{(x, Y) \sim \mathcal{D}}
\left[
\log 
\prod_{i=1}^{T'}
\frac{\exp(- E_{1}^{*}(x, y^{w}_{\leq i}))}{\sum_{y_{\leq i} \in Y_{i}} \exp(- E_{1}^{*}(x, y_{\leq i}))}
\right]\\
&=
- \mathbb{E}_{(x, Y) \sim \mathcal{D}}
\left[
\sum_{i=1}^{T'}
\log 
\frac{\exp(- E_{1}^{*}(x, y^{w}_{\leq i}))}{\sum_{y_{\leq i} \in Y_{i}} \exp(- E_{1}^{*}(x, y_{\leq i}))}
\right]\\
&=
- \mathbb{E}_{(x, Y) \sim \mathcal{D}}
\left[
\sum_{i=1}^{T'}
\log 
\frac{1}{1 + \exp(E_{1}^{*}(x, y^{w}_{\leq i}) - E_{1}^{*}(x, y^{l}_{\leq i}))}
\right]\\
&=
- \mathbb{E}_{(x, Y) \sim \mathcal{D}}
\left[
\sum_{i=1}^{T'}
\log \sigma(- (E_{1}^{*}(x, y^{w}_{\leq i}) - E_{1}^{*}(x, y^{l}_{\leq i})))
\right].\label{eq:a5}
\end{align}
where $T' = \max \{\mu'(y^{w}), \mu'(y^{l})\}$.
Taking the logarithm of both sides of Eq.~(\ref{eq:a_blz2}), we have
\begin{align}
\log p_{2}(y_{i}|y_{<i}, x) =
- E_{2}^{*}(x, y_{\leq i})
- \log \sum_{y_{i} \in \mathcal{V}} \exp(- E_{2}^{*}(x, y_{\leq i})),
\end{align}
and thus we have
\begin{align}
E_{2}^{*}(x, y_{\leq i})
=
- \log p_{2}(y_{i}|y_{<i}, x)
- \log \sum_{y_{i} \in \mathcal{V}} \exp(- E_{2}^{*}(x, y_{\leq i})).
\end{align}
From the definition of prefix energies, we obtain
\begin{align}
E_{1}^{*}(x, y_{\leq i})
&=
\beta \left( E_{2}^{*}(x, y_{\leq i}) + \log \pi_{\mathrm{ref}} (y_{i}|y_{<i}, x) \right)\\
&=
\beta \left( - \log p_{2}(y_{i}|y_{<i}, x)
- \log\hspace{-3pt}\sum_{y_{i} \in \mathcal{V}}\hspace{-3pt}\exp(- E_{2}^{*}(x, y_{\leq i})) + \log \pi_{\mathrm{ref}} (y_{i}|y_{<i}, x) \right) \hspace{-10pt}\\
&=
\beta \left( - \log \frac{p_{2}(y_{i}|y_{<i}, x)}{\pi_{\mathrm{ref}} (y_{i}|y_{<i}, x)}
 \right).
\end{align}
Thus, we have
\begin{align}
E_{1}^{*}(x, y^{w}_{\leq i}) - E_{1}^{*}(x, y^{l}_{\leq i})
&=
- \beta \log \frac{p_{2}(y^{w}_{i}|y^{w}_{<i}, x)}{\pi_{\mathrm{ref}} (y^{w}_{i}|y^{w}_{<i}, x)}
+ \beta \log \frac{p_{2}(y^{l}_{i}|y^{l}_{<i}, x)}{\pi_{\mathrm{ref}} (y^{l}_{i}|y^{l}_{<i}, x)}.
\label{eq:a10}
\end{align}

From Eqs.~(\ref{eq:a5}) and (\ref{eq:a10}), we obtain
\begin{align}
\mathcal{L}_{\text{ADPO}}
&=
- \mathbb{E}_{(x, Y) \sim \mathcal{D}}
\left[
\sum_{i=1}^{T'}
\log \sigma\left( \beta \log \frac{p_{2}(y^{w}_{i}|y^{w}_{<i}, x)}{\pi_{\mathrm{ref}} (y^{w}_{i}|y^{w}_{<i}, x)}
- \beta \log \frac{p_{2}(y^{l}_{i}|y^{l}_{<i}, x)}{\pi_{\mathrm{ref}} (y^{l}_{i}|y^{l}_{<i}, x)} \right)
\right].
\end{align}
When $p_{2} = \pi_{\theta}$, we have
\begin{align}
\mathcal{L}_{\text{ADPO}}
&=
- \mathbb{E}_{(x, Y) \sim \mathcal{D}}
\left[
\sum_{i=1}^{T'}
\log \sigma\left( \beta \log \frac{\pi_{\theta}(y^{w}_{i}|y^{w}_{<i}, x)}{\pi_{\mathrm{ref}} (y^{w}_{i}|y^{w}_{<i}, x)}
- \beta \log \frac{\pi_{\theta}(y^{l}_{i}|y^{l}_{<i}, x)}{\pi_{\mathrm{ref}} (y^{l}_{i}|y^{l}_{<i}, x)} \right)
\right].
\end{align}
\end{proof}

\section{KL-constrained Reward Maximization}
\label{app:klmax}

ADPO leverages the optimal solution of the KL-constrained reward maximization problem.
We first review Remark 1 ~\citep{Peters2007reinforcement, peng2019advantage, Korbak2022reinforcement, go2023aligning, rafailov2023DPO}, and then show Corollary 2.

\noindent\textit{\textbf{Remark 1 (KL-constrained Reward Maximization).}
Given a reward function $r(x, y)$, a reference model $\pi_{\mathrm{ref}}(y|x)>0$ and a constant $\beta>0$, we define the objective function for the KL-constrained reward maximization problem as
\begin{align}
\label{eq:klmax}
\mathcal{J}(\pi)
=
\mathbb{E}_{x \sim \mathcal{D}, y \sim \pi(y | x)} \bigl[ r(x, y) \bigr] - \beta D_{\textrm{KL}}\bigl[\pi(y | x) || \pi_{\mathrm{ref}}( y | x)\bigr].
\end{align}
Then, the optimal solution is given by the Boltzmann distribution induced by the posterior energy:
\begin{align}
E_{2}(x, y) = - \frac{1}{\beta} r(x, y) - \log \pi_{\mathrm{ref}} (y|x).
\end{align}
}

\begin{proof}
The Boltzmann distribution $p_{2}$ induced by $E_{2}$ is given by
\begin{align}
p_{2}(y|x) = \frac{\exp(-E_{2}(x,y))}{Z(x)}, \quad Z(x) = \sum_{y\in\mathcal{Y}} \exp(-E_{2}(x,y)).
\end{align}
Thus, we have
\begin{align}
\mathcal{J}(\pi)
&=
\mathbb{E}_{x \sim \mathcal{D}, y \sim \pi(y | x)} \bigl[ r(x, y) \bigr] - \beta D_{\textrm{KL}}\bigl[\pi(y | x) | | \pi_{\mathrm{ref}}( y | x)\bigr]\\
&=
\mathbb{E}_{x \sim \mathcal{D}, y \sim \pi(y | x)} \left[ r(x, y)  - \beta
\log \frac{\pi(y | x)}{\pi_{\mathrm{ref}}( y | x)} \right]\\
&=
- \beta \; \mathbb{E}_{x \sim \mathcal{D}, y \sim \pi(y | x)} \left[
\log \exp \left(- \frac{1}{\beta} r(x, y) \right) +
\log \frac{\pi(y | x)}{\pi_{\mathrm{ref}}( y | x)} \right]\\
&=
- \beta \; \mathbb{E}_{x \sim \mathcal{D}, y \sim \pi(y | x)} \left[
\log \frac{\pi(y | x)}{\pi_{\mathrm{ref}}( y | x) \exp \left(\frac{1}{\beta} r(x, y) \right)} \right]\\
&=
- \beta \; \mathbb{E}_{x \sim \mathcal{D}, y \sim \pi(y | x)} \left[
\log \frac{\pi(y | x)}{\exp\left(-E_{2}(x, y) \right)} \right]\\
&=
- \beta \; \mathbb{E}_{x \sim \mathcal{D}, y \sim \pi(y | x)} \left[
\log \frac{\pi(y | x)}{p_{2}(y|x)} - \log Z(x)\right]\\
&=
- \beta \; \mathbb{E}_{x \sim \mathcal{D}} \left[
D_{\mathrm{KL}}
\left(
\pi(y | x)||p_{2}(y|x)
\right) - \log Z(x)\right]
\end{align}
Then, we obtain
\begin{align}
\argmax_{\pi} \mathcal{J} (\pi)
=
\argmin_{\pi} D_{\mathrm{KL}}
\left(
\pi(y | x)||p_{2}(y|x)
\right)
= p_{2}
\end{align}
\end{proof}

\noindent\textit{\textbf{Corollary~2.}
Given a reward function $r(x,y)$ and a reference autoregressive model $\pi_{\mathrm{ref}}(y_{i}|y_{<i}, x) > 0$ and a constant $\beta > 0$, the following prefix posterior energy induces the optimal solution of the KL-constrained reward maximization problem in Eq.~(\ref{eq:klmax}):
\begin{align}
E_{2}^{*}(x, y_{\leq i}) = - \frac{1}{\beta} r^{*}(x, y_{\leq i}) - \log \pi_{\mathrm{ref}} (y_{i}|y_{<i}, x),
\end{align}
where $r^{*}$ is an additive decomposition of $r$.
}
\begin{proof}
Define an energy $\bar{E}^{*}_{2}$ as
\begin{align}
\bar{E}^{*}_{2}(x,y) = \sum_{i=1}^{T'} E_{2}^{*}(x, y_{\leq i})
\end{align}
Then, we have
\begin{align}
\bar{E}^{*}_{2}(x, y)
&= - \frac{1}{\beta} \sum_{i=1}^{T'} r^{*}(x, y_{\leq i}) -  \log  \prod_{i=1}^{T'} \pi_{\mathrm{ref}} (y_{i}|y_{<i}, x)\\
&= - \frac{1}{\beta} r(x, y) - \log  \pi_{\mathrm{ref}} (y|x)\\
&= E_{2}(x,y).
\end{align}
Therefore, we have $\bar{E}^{*}_{2} = E_{2}$.
From Remark 1, the Boltzmann distribution induced by $\bar{E}^{*}_{2}$ is the optimal solution of the KL-constrained reward maximization problem.
\end{proof}

\section{Prefix-wise Reparameterization Completeness}
\label{app:completeness}
\textit{\textbf{Proposition 1 (Prefix-wise Reparameterization Completeness).}
Let $[r^{*}]$ denote the reward-shift equivalence class of a prefix-wise reward function $r^{*} : \mathcal{X} \times \mathcal{Y}^{*} \to \mathbb{R}$, defined as
\begin{align}
[r^{*}] = \{ r' \;|\;
\exists f \;
\forall (x, y_{\leq i}) \;
r'(x,y_{\leq i}) = r^{*}(x,y_{\leq i}) + f(x,y_{<i})\}.
\end{align}
Given an autoregressive reference model $\pi_{\mathrm{ref}}(y_{i}|y_{<i},x) > 0$ and a hyperparameter $\beta > 0$,
for any prefix-wise reward function $r^{*}$,
there exists a unique representative $r^{*}_{\circ} \in [ r^{*} ]$ such that, for all $x \in \mathcal{X}$ and prefixes $y_{\leq i} \in \mathcal{Y}^{*}$,
\begin{align}
\label{eq:representative_circ}
r^{*}_{\circ}(x, y_{\leq i})
\equiv
\beta \log
\frac{\pi (y_{i} | y_{<i}, x)}{\pi_{\mathrm{ref}} (y_{i} | y_{<i}, x)}.
\end{align}
for some \uline{autoregressive} model $\pi$.}
\begin{proof}
Given a prefix-wise reward function $r^{*}$, we first show the existence of $\pi$.
Define the Boltzmann distribution
\begin{align}
\pi(y_{i}| y_{<i},x)
=
\frac{\exp(- E_{2}^{*}(x, y_{\leq i}))}{Z(x, y_{<i})},
\quad
Z(x, y_{<i})
=
\sum_{y_{i}}
\exp(- E_{2}^{*}(x, y_{\leq i}))
\end{align}
Taking logarithms gives
\begin{align}
\log \pi(y_{i}| y_{<i},x)
&=
- E_{2}^{*}(x, y_{\leq i})
- \log Z(x, y_{<i})\\
&= \frac{1}{\beta} r^{*}(x, y_{\leq i}) + \log \pi_{\mathrm{ref}}(y_{i}| y_{<i},x)- \log Z(x, y_{<i})
\end{align}
Thus, we obtain the required form:
\begin{align}
r^{*}(x, y_{\leq i})
=
\beta \log \frac{\pi(y_{i}| y_{<i},x)}{\pi_{\mathrm{ref}}(y_{i}| y_{<i},x)}
+\beta \log Z(x, y_{<i}).
\end{align}
If another pair $(\tilde r, \tilde\pi)$ enjoys the same property, their
log-ratios differ by a function independent of~$y_i$, thereby $\tilde r-r^{*}_{\circ}=f(x,y_{<i})$.
Hence both belong to the same shift class, $r^{*}_{\circ}$ is unique modulo such shifts.
\end{proof}

\section{Evaluation Metrics}
\label{app:eval-metrics}

In this section, we provide detailed descriptions of the evaluation metrics used in our experiments.

\subsection{Math Reasoning Tasks}

For GSM8K and MATH500, we report \textbf{accuracy}.  
A prediction is counted as correct if the final numerical answer matches the ground-truth answer after standard normalization (e.g., stripping formatting, ignoring trivial differences).  

\subsection{Conversational Benchmarks}

\paragraph{AlpacaEval 2}
is an automatic preference-evaluation benchmark that uses an LLM-as-a-judge framework to assess instruction-following quality.  
We report two metrics:
\begin{itemize}
    \item \textbf{Win Rate (WR):} the percentage of pairwise comparisons in which the model's response is preferred over the baseline.
    \item \textbf{Length-Controlled Win Rate (LC):} a debiased variant of WR that adjusts for response-length bias.
\end{itemize}

\paragraph{Arena-Hard}
is a curated set of difficult prompts from the Chatbot Arena, designed to measure robustness on challenging instruction-following tasks.  
We report:
\begin{itemize}
    \item \textbf{Win Rate:} the fraction of pairwise comparisons in which the model's response is preferred according to the Arena-Hard judging protocol (LLM-based judges).
\end{itemize}

\paragraph{MT-Bench}
is a multi-turn dialogue benchmark consisting of 80 conversational questions spanning various categories (e.g., reasoning, coding, safety).  
Evaluation is conducted with GPT-4-based judges.  
We report:
\begin{itemize}
    \item \textbf{MT-Bench Score:} the average judge rating on a 1--10 scale across all turns, reflecting helpfulness, coherence, and overall response quality.
\end{itemize}

\section{Additional Experiments}
\label{app:exp}
In this section, we conduct additional experiments to investigate the impact of several hyperparameters, including the hyperparameter $\beta$ in the objective function, LoRA rank, and model size. We also analyze the learning behavior in the conversational task. The results collectively validate the consistent effectiveness of ADPO across diverse experimental settings and model configurations.

\paragraph{Hyperparameter $\beta$.} 
Table~\ref{tab:beta} provides a hyperparameter analysis of $\beta$ for DPO and ADPO across four LLMs evaluated on GSM8K. ADPO consistently surpasses DPO, demonstrating notable improvements particularly for smaller $\beta$ values (0.5 and 1.0).

\paragraph{LoRA Rank.}
Table~\ref{tab:lora_rank} compares the effects of different LoRA ranks (4, 16, and 64) on GSM8K for Llama-3-8B.
We observe that ADPO outperforms DPO across all ranks and the performance gap widens at higher ranks.

\paragraph{Model Size.}
Table~\ref{tab:model_size} examines the impact of scaling model size from Llama-3-8B to Llama-3-70B on GSM8K.
ADPO consistently outperforms DPO at both model sizes, achieving 89.23\% with a size of 70B.

\begin{table}[h]
\centering
\setlength{\tabcolsep}{3.6pt}
\caption{Hyperparameter study for $\beta$. Results are reported for $\beta = 0.5, 1.0, 1.5$ on GSM8K.}
\begin{tabular}{lcccccccccccc}
\toprule
\multirow{2}{*}{\textbf{Method}} 
& \multicolumn{3}{c}{\textbf{Llama-3-8B}} 
& \multicolumn{3}{c}{\textbf{Gemma-3-12B}} 
& \multicolumn{3}{c}{\textbf{Qwen-3-8B}} 
& \multicolumn{3}{c}{\textbf{DeepSeek-Math-7B}} \\[0.1em]
\cmidrule(lr){2-4} \cmidrule(lr){5-7} \cmidrule(lr){8-10} \cmidrule(lr){11-13}
& $0.5$ & $1.0$ & $1.5$ 
& $0.5$ & $1.0$ & $1.5$ 
& $0.5$ & $1.0$ & $1.5$ 
& $0.5$ & $1.0$ & $1.5$ \\
\midrule
DPO &
62.47 & 64.37 & 63.91 &
77.71 & 77.03 & 75.82 &
87.87 & 86.96 & 86.73 &
66.94 & 67.78 & 67.63 \\
ADPO &
68.61 & 68.08 & 64.90 &
78.85 & 78.32 & 77.48 & 
88.10 & 88.10 & 88.70 &
70.05 & 69.98 & 69.22 \\
\bottomrule
\end{tabular}
\label{tab:beta}
\end{table}

\begin{table}[h]
\centering
\begin{minipage}{0.45\linewidth}
\centering
\caption{LoRA rank $r$ (Llama-3-8B).}
\label{tab:lora_rank}
\setlength{\tabcolsep}{7pt}
\begin{tabular}{lccc}
\toprule
\textbf{Method} & 4 & 16 & 64\\
\midrule
DPO & 63.53 & 64.37 & 63.84\\
ADPO & 65.81 & 68.08 & 68.23\\
\bottomrule
\end{tabular}
\end{minipage}
\begin{minipage}{0.45\linewidth}
\centering
\caption{Model Size.}
\label{tab:model_size}
\begin{tabular}{lcc}
\toprule
\textbf{Method} & Llama-3-8B & Llama-3-70B \\
\midrule
DPO & 64.37 & 88.10\\
ADPO & 68.08 & 89.23\\
\bottomrule
\end{tabular}
\end{minipage}
\end{table}

\begin{table}[t]
\centering
\caption{Full-parameter tuning results on Llama-3-8B (training without LoRA).}
\label{tab:full_parameter}
\begin{tabular}{lcc}
\toprule
\textbf{Method} & GSM8K & MATH \\
\midrule
DPO & 64.37 & 18.00 \\
\rowcolor{gray!12}
ADPO & 67.32 & 20.00 \\
\midrule
cDPO & 66.19 & 18.00 \\
\rowcolor{gray!12}
cADPO & 68.39 & 19.80 \\
\bottomrule
\end{tabular}
\end{table}

\paragraph{Full parameter tuning.}
Table~\ref{tab:full_parameter} provides the results of full-parameter tuning (\textit{i}.\textit{e}. without LoRA) for Llama-3-8B.
ADPO and cADPO still outperform DPO and cDPO even under full-parameter training.

\begin{table}[t]
\setlength{\tabcolsep}{3.2pt}
\small
\centering
\caption{Standard deviations on two mathematical reasoning benchmarks.}
\begin{tabular}{lcccccccc}
\toprule
\multirow{2}{*}{\textbf{Method}} 
& \multicolumn{2}{c}{\textbf{Llama-3-8B}} 
& \multicolumn{2}{c}{\textbf{Gemma-3-12B}} 
& \multicolumn{2}{c}{\textbf{Qwen-3-8B}} 
& \multicolumn{2}{c}{\textbf{DeepSeek-Math-7B}} \\[0.1em]
\cmidrule(lr){2-3} \cmidrule(lr){4-5} \cmidrule(lr){6-7} \cmidrule(lr){8-9}
& GSM8K & MATH 
& GSM8K & MATH 
& GSM8K & MATH 
& GSM8K & MATH \\
\midrule
DPO
& 0.19 & 0.33 & 0.22 & 0.20 & 0.18 & 0.61 & 0.24 & 0.45
\\[0.1em]
\rowcolor{gray!12}
ADPO &
0.19 & 0.33 & 0.22 & 0.32 & 0.06 & 0.73 & 0.29 & 0.41\\
\midrule
cDPO &
0.09 & 0.32 & 0.14 & 0.15 & 0.12 & 0.20 & 0.21 & 0.43 \\[0.1em]
\rowcolor{gray!12}
cADPO
& 0.08 & 0.31 & 0.17 & 0.47 & 0.13 & 0.45 & 0.11 & 0.41 \\
\bottomrule
\end{tabular}
\label{tab:math_std}
\end{table}

\begin{table}[t]
\centering
\caption{Standard errors of win rates and scores on three conversational benchmarks.}
\begin{tabular}{lcccc}
\toprule
\multirow{3}{*}{\raisebox{-8pt}{\textbf{Method}}} 
& \multicolumn{4}{c}{\textbf{Llama3-8B}} \\
\cmidrule(lr){2-5}
& \multicolumn{2}{c}{\textbf{AlpacaEval~2}} & \textbf{Arena-Hard} & \textbf{MT-Bench} \\
\cmidrule(lr){2-3} \cmidrule(lr){4-4} \cmidrule(lr){5-5} 
& LC & WR & WR & Score \\
\midrule
SFT & 0.75 & 1.28 & 1.5 & 0.08 \\
DPO & 0.81 & 1.46 & 1.4 & 0.08\\
SimPO & 0.80 & 1.47 & 1.4 & 0.08\\
\rowcolor{gray!12}
ADPO (Ours) & 0.79 & 1.49 & 1.4 & 0.09\\
\bottomrule
\end{tabular}
\label{tab:conv_std}
\end{table}

\begin{table}[t]
\setlength{\tabcolsep}{3.2pt}
\small
\centering
\caption{Average response length.}
\begin{tabular}{lcccccccc}
\toprule
\multirow{2}{*}{\textbf{Method}} 
& \multicolumn{2}{c}{\textbf{Llama-3-8B}} 
& \multicolumn{2}{c}{\textbf{Gemma-3-12B}} 
& \multicolumn{2}{c}{\textbf{Qwen-3-8B}} 
& \multicolumn{2}{c}{\textbf{DeepSeek-Math-7B}} \\[0.1em]
\cmidrule(lr){2-3} \cmidrule(lr){4-5} \cmidrule(lr){6-7} \cmidrule(lr){8-9}
& GSM8K & MATH 
& GSM8K & MATH 
& GSM8K & MATH 
& GSM8K & MATH \\
\midrule
DPO
& 317.8 & 869.7 & 391.2 & 895.2 & 299.7 & 513.8 & 257.7 & 682.6
\\[0.1em]
\rowcolor{gray!12}
ADPO (Ours) &
342.8 & 917.9 & 518.5 & 2192.3 & 294.9 & 525.7 & 280.8 & 662.9 \\
\midrule
cDPO &
520.4 & 1859.7 & 333.5 & 1174.9 & 299.8 & 526.9 & 261.7 & 696.7 \\[0.1em]
\rowcolor{gray!12}
cADPO (Ours)
& 308.1 & 847.0 & 287.2 & 841.0 & 313.8 & 510.4 & 250.3 & 630.0 \\
\bottomrule
\end{tabular}
\label{tab:length}
\end{table}

\begin{table}[t]
\centering
\caption{Wall-clock training time on GSM8K with Llama-3-8B.}
\label{tab:training_time}
\begin{tabular}{lc}
\toprule
\textbf{Method} & \textbf{Time (seconds)} \\
\midrule
DPO & 2202 \\
\midrule
ADPO ($\xi_{\text{static}}$, $k=4$) & 2314 \\
ADPO ($\xi_{\text{static}}$, $k=2$) & 2228 \\
ADPO ($\xi_{\text{static}}$, $k=1$) & 2301 \\
\midrule
ADPO ($\xi_{\text{adaptive}}$, $m=1$)$^{\dagger}$ & 2234 \\
ADPO ($\xi_{\text{adaptive}}$, $m=2$) & 2287 \\
ADPO ($\xi_{\text{adaptive}}$, $m=16$) & 2321 \\
ADPO ($\xi_{\text{adaptive}}$, $m=128$) & 2318 \\
ADPO ($\xi_{\text{adaptive}}$, $m=256$) & 2226 \\
ADPO ($\xi_{\text{adaptive}}$, $m=512$) & 2327 \\
\bottomrule
\multicolumn{2}{l}{\footnotesize $^{\dagger}$Equivalent to DPO.}
\end{tabular}
\end{table}

\paragraph{Standard Deviation.}
Table~\ref{tab:math_std} reports the standard deviations over three independent training runs corresponding to the results in Table~\ref{tab:math} of the main paper.
Table~\ref{tab:conv_std} reports the standard errors of the win rates and scores obtained via LLM-as-a-Judge corresponding to the results in Table~\ref{tab:conversation} of the main paper.
We observe that the standard deviations (mathematical tasks) and standard errors (conversational tasks) are consistently small. The improvements are consistently larger than the observed errors, and the relative ranking of methods remains unchanged across all settings, supporting the robustness of our conclusions.

\paragraph{Generation Length.}
Table~\ref{tab:length} reports the average output length across all models and datasets.
We observed two trends: (a) ADPO produces slightly longer responses than DPO, while (b) cADPO produces shorter responses than cDPO.

ADPO performs preference comparison at each prefix, encouraging the model to accumulate positive evidence step-by-step.
This finer-grained signal encourages the model to articulate intermediate reasoning more explicitly, which naturally results in slightly longer outputs.

cDPO amplifies penalties for ``critical tokens'' at the sequence level.
This can lead to longer outputs, because producing additional reasoning steps gives the model more chances to place tokens that are not penalized.
In contrast, cADPO applies the same penalties locally at each prefix, preventing such length-driven attenuation.
As a result, cADPO discourages unnecessary elaboration and yields more concise outputs.

\begin{figure}[t]
    \centering
    \includegraphics[width=\linewidth]{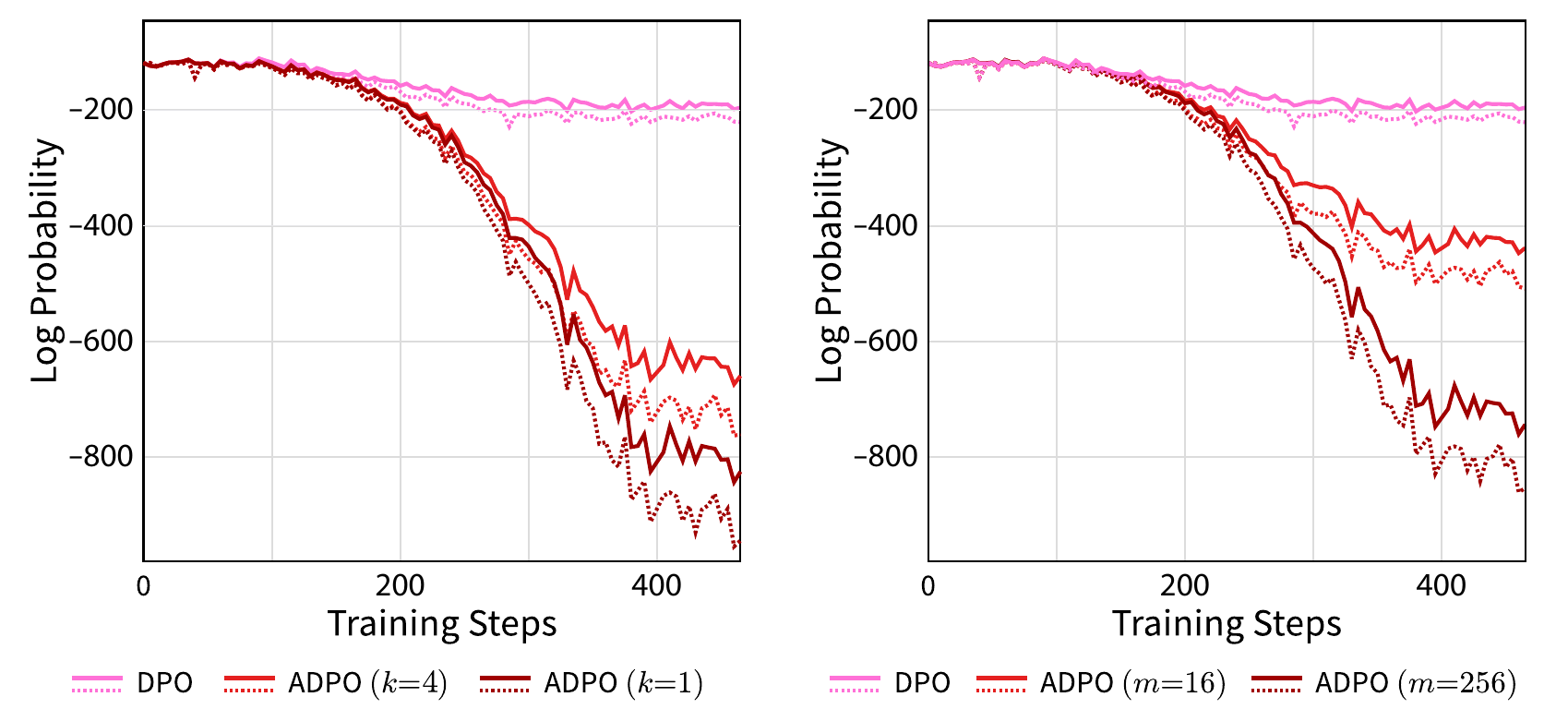}
    \caption{Training dynamics (conversation task).}
    \label{fig:adpo_plot_conv}
\end{figure}

\paragraph{Computational Complexity.}
ADPO introduces additional prefix-wise computations compared to DPO, which may raise concerns about increased training cost.
To assess this, we measured the wall-clock training time on GSM8K with Llama-3-8B across all ADPO configurations and compared them with DPO.
As reported in Table~\ref{tab:training_time}, the difference between DPO and the slowest ADPO variant is 125 seconds (5.7\%), indicating that the additional computational overhead of ADPO is negligible in practice.

\paragraph{Learning Behavior.} Figure~\ref{fig:adpo_plot_conv} shows training dynamics for the conversation task by plotting the evolution of sequence log probabilities over training steps for DPO and ADPO.
In the static family (left), reducing the window size increases the margin between preferred (solid lines) and dispreferred (dashed lines) trajectories, showing greater separation compared to DPO.
In the adaptive family (right), increasing the number of segments similarly amplifies this margin. Across both settings, ADPO consistently achieves a clearer divergence between preferred and dispreferred responses throughout training.

\paragraph{Implementation Details.}
We implemented our methods using the Transformer Reinforcement Learning (TRL) library. For training on GSM8K, we followed the experimental setup of \cite{lin2025critical}. Specifically, positive and negative models are first trained for 1 epoch with a learning rate of $3\!\times\!10^{-4}$. Subsequently, preference optimization models are trained with $\beta=1.0$ for 3 epochs using AdamW with LoRA ($r=16$), at a learning rate of $2\!\times\!10^{-5}$ for DPO and $4\!\times\!10^{-5}$ for cDPO. Each problem was sampled 64 times with a top-p probability of 50\% during contrastive estimation.
For the conversation task, all models are trained for 1 epoch using preference data scored by PairRM. Experiments are conducted using four NVIDIA H100 GPUs. We will release our code and models.

\section{Additional Analysis}

\subsection{Prefix-wise Reward Dynamics}

\begin{figure}[t]
\centering
\includegraphics[width=\linewidth]{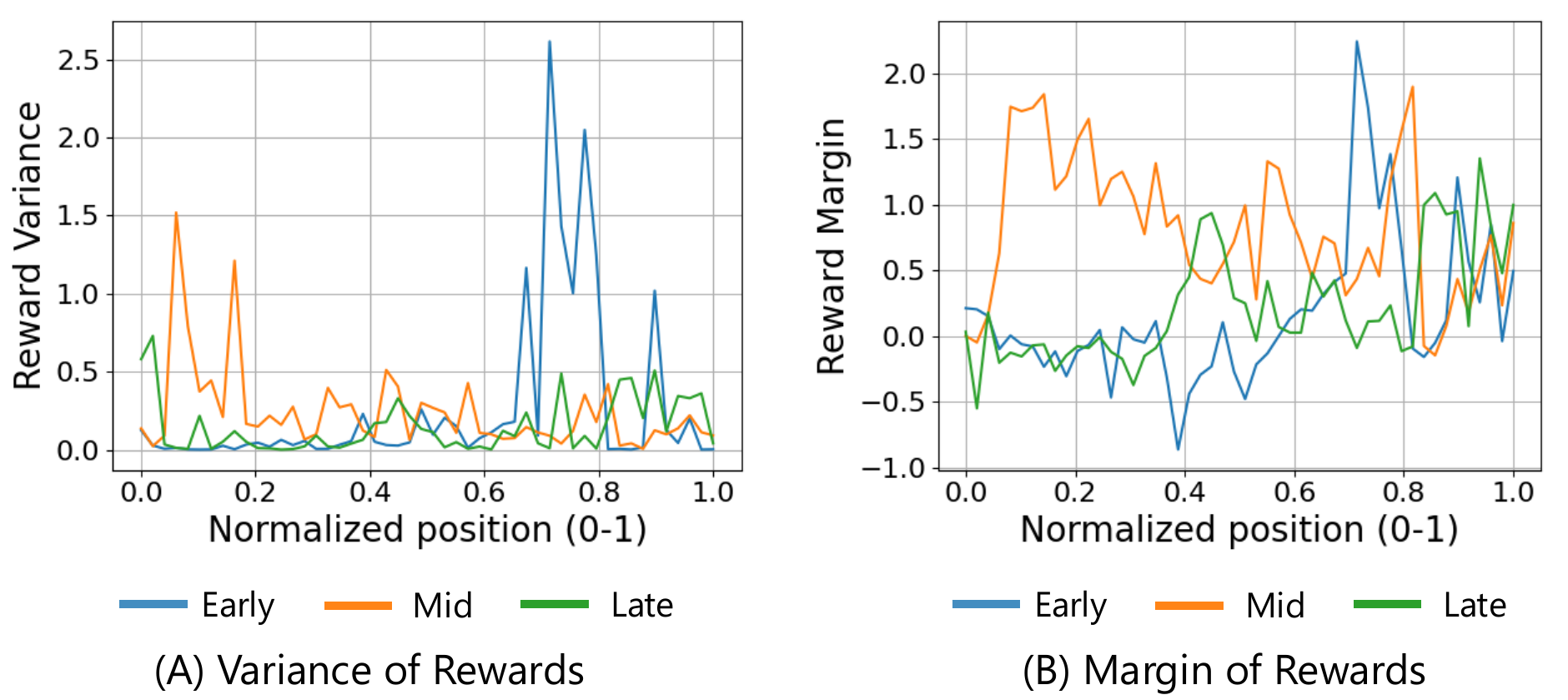}
\caption{Prefix-wise reward variance and margin.}
\label{fig:prefix_reward_dynamics}
\end{figure}

As formalized in Eq.~\ref{eq:representative}, ADPO assigns an implicit reward at each step $i$.
To understand how these implicit rewards are distributed across prefixes and how they evolve during training, we analyze two summary statistics of the prefix-wise reward signal: 
(A) the reward variance, which reflects how stable or uncertain the reward is at each prefix, and 
(B) the reward margin (chosen--rejected difference), which captures where preference differences are realized.
We normalize prefix positions to $[0,1]$ and evaluate these quantities at early, middle, and late checkpoints from ADPO training on the Llama-3-8B model for the math reasoning task.
Figures~\ref{fig:prefix_reward_dynamics}~(A) and~(B) illustrate how the variance and margin profiles evolve across training.

\paragraph{Early Training.}
As shown in the blue curves in Figures~\ref{fig:prefix_reward_dynamics} (A) and (B), both reward variance and margin are concentrated in the middle-to-late prefix region.
This indicates that ADPO first focuses on correcting the unstable reasoning steps that occur
near the end of the response, where chosen and rejected outputs differ most 
and the policy is initially most uncertain.

\paragraph{Middle Training.}
In the mid-stage checkpoints (orange curves), the mass of both variance and margin
shifts toward earlier prefixes.
Once the later reasoning stabilizes, ADPO begins refining the early parts of the reasoning chain: these early steps determine how the model sets up the solution and whether the subsequent reasoning flows toward a correct final answer.

\paragraph{Late Training.}
In the late-stage checkpoints (green curves), reward variance becomes small across all prefixes, while the margin becomes sharply concentrated in the final portion of the sequence.
This suggests that the global reasoning structure has largely converged, and ADPO focuses on 
fine-tuning the last few steps that directly determine correctness.

\paragraph{Summary.}
Overall, these results reveal a consistent three-phase progression: 
ADPO first improves the end of the reasoning path, then the beginning, and finally sharpens the answer-producing tokens themselves.
This demonstrates that ADPO performs meaningful prefix-wise credit assignment, progressively localizing the implicit reward to the prefixes that matter most for preference.

\end{document}